\documentclass[onecolumn,11pt]{article}

\usepackage[top=1in, bottom=1in, left=1in, right=1in]{geometry}
\usepackage[numbers]{natbib}

\usepackage[colorlinks=true]{hyperref}  
\hypersetup{allcolors = blue}           
\usepackage{amsthm}
\usepackage{amssymb,bm}
\usepackage{amsmath}  
\usepackage{url}
\usepackage{indentfirst}                
\usepackage{pxfonts}
\usepackage[T1]{fontenc}                

\usepackage{graphicx}
\usepackage{subfigure}
\usepackage{setspace}
\usepackage{color}

\usepackage{arydshln}

\usepackage{supertabular}
\usepackage{booktabs}

\usepackage[section]{placeins}
\usepackage{floatrow}
\usepackage[noend]{algpseudocode}
\usepackage[utf8]{inputenc}
\usepackage{fontenc}
\usepackage{lipsum}
\usepackage[ruled,linesnumbered]{algorithm2e}

\newtheorem{theorem}{Theorem} 
\newtheorem{definition}{Definition} 
\newtheorem{proposition}{Proposition} 

\newcommand{\RNum}[1]{\uppercase\expandafter{\romannumeral #1\relax}}

\graphicspath{{figs/}}
\newcommand{\RR}{\mathbb{R}}

\newcommand{\N}{\mathbb{N}}
\usepackage{tikz}
\usepackage{overpic}

\usepackage[utf8]{inputenc}
\usepackage{setspace}

\renewcommand{\abstractname}{\vspace{-\baselineskip}}   

\newcommand\blfootnote[1]{%
	\begingroup
	\renewcommand\thefootnote{}\footnote{#1}%
	\addtocounter{footnote}{-1}%
	\endgroup
}

\title{\vspace{-.4in}{\fontsize{15}{15}\selectfont \textbf{Learning Dynamical Systems from Data: A Simple Cross-Validation Perspective, Part V: Sparse Kernel Flows for 132 Chaotic Dynamical Systems}
	}}

\author{\normalsize{Lu Yang$^{1,4}$, Xiuwen Sun$^{5}$, Boumediene Hamzi$^{2,3*}$, Houman Owhadi$^2$, Naiming Xie$^1$}\\
	\footnotesize{$^1$ College of Economics and Management, Nanjing University of Aeronautics and Astronautics, PR China}\\
	\footnotesize{$^2$ Department of Computing and Mathematical Sciences, Caltech, CA, USA}\\
	\footnotesize{$^3$ Department of Applied Mathematics and Statistics, Johns Hopkins University, Baltimore, MD, USA.}\\
	\footnotesize{$^4$ Department of Mathematics, Imperial College London, United Kingdom}\\
	\footnotesize{$^5$ Zhejiang Zhelixin Credit Investigation Co., Ltd. \vspace{-.2in}}
}
\date{}

\begin{document}

\maketitle

\blfootnote{$^*$ Corresponding author (boumediene.hamzi@gmail.com).}

\vspace{-.2in}

\setlength\baselineskip{15.3pt}

\begin{abstract} 
	\linespread{1}\selectfont
	\noindent\textbf{Abstract}~
	Regressing the vector field of a dynamical system from a finite number of observed states is a natural way to learn surrogate models for such systems. As shown in \cite{BHPhysicaD, bhkfnp, lee2021learning, bhkfsdes, hamzimaulikowhadi,bh_kfs_missing,bh_hamiltonian}, a simple and interpretable way to learn a dynamical system from data is to interpolate its vector-field with a data-adapted kernel which can be learned by using Kernel Flows \cite{Owhadi19}. 
	
	The method of Kernel Flows is a trainable machine learning method that learns the optimal parameters of a kernel based on the premise that a kernel is good if there is no significant loss in accuracy if half of the data is used. The objective function could be a short-term prediction or some other objective (cf. \cite{BHPhysicaD} and \cite{Lu_Hausdorf} for other variants of Kernel Flows). However, this method is limited by the choice of the base kernel. 
	
	In this paper, we introduce the method of \emph{Sparse Kernel Flows } in order to learn the ``best'' kernel by starting from a large dictionary of kernels. It is based on sparsifying a kernel that is a linear combination of elemental kernels. We apply this approach to a library of 132 chaotic systems.
	
	\vspace{6pt}
	

\end{abstract}

\section{Introduction}

The ubiquity of time series in many domains of science has led to the development of diverse statistical and machine learning forecasting methods \cite{kantz97,CASDAGLI1989, yk1, yk2, yk3, yk4, survey_kf_ann,Sindy,jaideep1,nielsen2019practical,abarbanel2012analysis}.

Amongst various learning-based approaches, kernel-based methods hold the potential for considerable advantages  in terms of theoretical
analysis, numerical implementation, regularization, guaranteed convergence, automatization, and interpretability \cite{chen2021solving, houman_cgc}. Indeed, reproducing kernel Hilbert spaces (RKHS) \cite{CuckerandSmale} has provided strong mathematical foundations for analyzing dynamical systems \cite{yk1, bhcm11,bhcm1,lyap_bh,bh2020a,hamzi2019kernel, bh2020b,klus2020data,ALEXANDER2020132520,bhks,bh12,bh17,hb17,mmd_kernels_bh} and surrogate modelling (cf. \cite{santinhaasdonk19} for a survey). Yet, the accuracy of these emulators depends on the kernel, but limited attention has been paid to the problem of selecting a good kernel.

Recently, experiments by Hamzi and Owhadi \cite{Owhadi19} have shown that kernel flows (KFs, an RKHS technique) can successfully reconstruct the dynamics of prototypical chaotic dynamical systems under both regular and irregular \cite{lee2021learning} sampling in time. Given a parameterized kernel function, KFs utilizes the regression relative error between two interpolants represented by the kernel (one is obtained by all data points and another is obtained by halved data points) as the quantity to minimize. In this sense, it can also be viewed as a variant of the cross-validation method. Later on, several research has extended the regular Kfs. The nonparametric version of kernel flows, based on kernel warping, is used to approximate chaotic dynamical systems in \cite{bhkfnp}. A KFs version for SDEs is at \cite{bhkfsdes}. A version for systems with missing dynamics is at \cite{bh_kfs_missing} and for Hamiltonian dynamics is at \cite{bh_hamiltonian}. Another version based on the Haudorff distance to learn attractors is at \cite{Lu_Hausdorf}.  From the application perspective, it was used in a machine learning context for classification \cite{yoo2021deep} and more recently in geophysical forecasting \cite{hamzi2022simple}.

It is worth noting that although the aforementioned kernel learning algorithms can obtain optimal parameters of the kernel from data, it requires a base kernel, making it sub-optimal for practitioners to select an appropriate kernel function for a given practical problem. In this work, we introduce the method of \emph{Sparse Kernel Flows} to discover the 'best' kernel from time series. It is based on starting with a base kernel that contains a large number of terms and then sparsifying it by setting some of the initial terms to zero. 
The principal contributions are summarized as follows:
\begin{enumerate}
	\item[(1)] Given a base kernel that consists of the sum of a relatively large number of elemental kernels, we develop the algorithm of \emph{Sparse Kernel Flows}. The sparsity is obtained via $\ell_1$-regularization. Starting from the initial base kernel, the goal of the method is to seek the ``best'' kernel with as few terms as possible  with the ``best'' emulation characteristic. 
	\item[(2)] In order to evaluate the generalization performance and accuracy of our proposed Sparse Kernel Flows algorithm, we apply the proposal to a chaotic dynamical system benchmark including 132 systems.
\end{enumerate}

The remaining parts are organized as follows. Section 2 states the problem and proposes the Sparse Kernel Flows algorithm. Section 3 consists of numerical experiments where we apply the method of Sparse Kernel Flows to the 132 chaotic dynamical systems library in \cite{gilpin2021chaos} to evaluate the generalization and accuracy. Section 4 concludes the work.

\section{Statement of the problem and its proposed method}
 
Given a time series $\bm{x}_1,\cdots, \bm{x}_n$ from a deterministic dynamical system in $ \mathbb{R}^d$, our goal is to forecast the evolution of the dynamics from historical observations.

A natural solution to forecasting the time series is to assume that the data are sampled from a discrete dynamical system
\begin{equation}
	\bm{x}_{k+1} = f^\dagger(\bm{x}_k,\cdots,\bm{x}_{k-\tau^\dagger +1})
\end{equation}
where $\bm{x}_k \in \mathbb{R}^d$ is the state of the system at time $t_k$, $f^\dagger$ represents the unknown vector field and $\tau^\dagger \in \mathbb{N}^*$ represents the delay embedding or delay. 

In order to approximate $f^\dagger$, given $\tau \in \mathbb{N}^*$ (selection strategies of $\tau$  are detailed in \cite{hamzi2021learning}), the problem of the dynamical system approximation can be recast as a kernel interpolation  problem (employed by \cite{owhadi2019kernel})
\begin{equation}
	\bm{Y}_k = f^\dagger(\bm{X}_k), ~k = 1,\cdots,N
\end{equation}
with $\bm{X}_k:=\left(\bm{x}_{k+\tau-1},\cdots, \bm{x}_k \right)$, $\bm{Y}_k:=\bm{x}_{k+\tau}$ and $N = n-\tau$ for $\tau \in \mathbb{N}^*$ . 

Given a reproducing kernel Hilbert space\footnote{A brief overview of RKHSs is given in the Appendix.} of candidates $\mathcal{H}$ for $f^\dagger$, and using the relative error in the RKHS norm $\|\cdot\|_\mathcal{H}$ as a loss, the regression of the data $(\bm{X}_k,\bm{Y}_k)$ with the kernel $k: \mathbb{R}^n \times  \mathbb{R}^n \rightarrow \mathbb{R}$ associated with $\mathcal{H}$ provides a minimax optimal approximation \cite{owhadi_scovel_2019}  of  $f^\dagger$ in $ \mathcal{H}$.  This regressor (in the presence of measurement noise of variance $\lambda>0$) is
\begin{equation}\label{mean_gp}
	f(\bm{x}) = K(\bm{x},\bm{X})\left(K(\bm{X},\bm{X}) + \lambda\bm{I} \right)^{-1}\bm{Y}
\end{equation}
where $\bm{X} = \left( \bm{X}_1, \cdots, \bm{X}_N \right), ~\bm{Y} = \left( \bm{Y}_1, \cdots, \bm{Y}_N \right) $, $K(\bm{x},\bm{X})$ is the $N \times N$ matrix with entries $k(\bm{x},\bm{X}_i)$, $K(\bm{X},\bm{X})$ is the $N \times N$ matrix with entries $k(\bm{X}_i,\bm{X}_j)$, $\bm{I}$ is the identity matrix and $\lambda_1\ge 0$ is a hyper-parameter that ensures the matrix $K(\bm{X},\bm{X}) + \lambda\bm{I}$ invertible. This regressor has also a natural interpretation in the setting of Gaussian process (GP) regression:  (\ref{mean_gp}) is the conditional mean of the centred GP $\xi \sim \mathcal{N}(0,K)$  with covariance function K conditioned on $\xi(\bm{X}_k) = \bm{Y}_k+\sqrt{\lambda}\bm{Z}_k$ where the $\bm{Z}_k$ are centred i.i.d. normal random variables. 

\subsection{A reminder on the Kernel Flows (KF) algorithm}\label{subseckf}
The accuracy of any kernel-based method depends on the kernel $k$.  Here we follow the parametrised KFs algorithm \cite{BHPhysicaD} to learn a "good" kernel in the sense that if the number of regression points can be halved without significant loss in accuracy (measured by $\|\cdot\|_\mathcal{H}$ associated with the kernel).

To describe this algorithm, let  $k_\theta (\bm{x},\bm{x}')$ be a family of kernels parameterized by $\bm{\theta}$,  and let $K_\theta (\bm{x},\bm{x}')$ be the corresponding gram matrix. The kernel learning iteration procedures of KFs can be described as follows

\begin{enumerate}
	\item[(1)] Consider a parametrized kernel function $k_\theta$ and initialize the parameters $\bm{\theta}^0$;
	\item[(2)] Prepare the data vector $\bm{X}^b = \left( \bm{X}_1,\cdots,\bm{X}_N \right) $ and  $\bm{Y}^b=\left( \bm{Y}_1,\cdots,\bm{Y}_N \right) $ according to the time delay $\tau$;
\end{enumerate}

Repeat the following until convergence:
\begin{enumerate}
	\item[(4)] Select $|\bm{X}^c| = \lfloor \frac{|\bm{X}^b|}{2} \rfloor$ and $|\bm{Y}^c| = \lfloor \frac{|\bm{Y}^b|}{2} \rfloor$ observations at random among the $N$ observations;
	\item[(5)] Design the loss function  
				\begin{equation}
					\begin{aligned}
						{\rho}(\bm{\theta}) &= 
						\frac{\|f^b-f^c\|^2_\mathcal{H}}{\|f^b\|^2_\mathcal{H}} 
						= 1 - \frac{{\bm{Y}^c}^\top\left(  K_{\theta}(\bm{X}^c,\bm{X}^c) +\lambda\bm{I}\right)^{-1} \bm{Y}^c}{{\bm{Y}^b}^\top \left( K_{\theta}(\bm{X}^b,\bm{X}^b)+\lambda\bm{I} \right) ^{-1} \bm{Y}^b}
					\end{aligned}
				\end{equation}
	which is the squared relative error (in the RKHS norm $\|\cdot\|_{k_\theta}$ defined by $k_\theta$) between the interpolants $f^b$ and $f^c$ obtained from the two nested subsets of the time series;
	\item[(6)] Compute the gradient $\nabla_\theta \rho$ with respect of the parameters $\bm{\theta}^k$;
	\item[(7)] Evolve the parameter in the gradient descent direction of $ \rho$: $\bm{\theta}^{k+1} \leftarrow \bm{\theta}^k - \delta\nabla_\theta \rho$.
\end{enumerate}

From the training procedures, a key observation is that designing an appropriate kernel function is essential for a specific system approximation problem. While this kernel learning method allows learning the parameters of a kernel, it also allows some nonzero terms to have very small magnitudes, which may lead to a loss of accuracy since they could be zero in reality. 

\subsection{Proposed Sparse Kernel Flows algorithm}
In this section we present the Sparse Kernel Flows algorithm that uses the Least absolute shrinkage and selection operator (Lasso) \cite{tibshirani1996regression} to sparsify the base kernel.  Without loss of generality, let us assume that the base kernel   can be written as a linear combination of $m$  kernels:
\begin{equation}
	k_{{\alpha},{\theta}} = \sum_{i=1}^{m} \bm{\alpha}^2_i k_i(\bm{x},\bm{x}^{\prime};\bm{\theta})
\end{equation}
where $k_i$ is a candidate kernel term, $\bm{\theta} = \left(\theta_1,\cdots,\theta_m \right)$ is the intrinsic parameter of candidate kernels, and $\bm{\alpha} = \left(\alpha_1,\cdots,\alpha_m \right) $ is the set of  weight coefficients that determine the degree of which kernels are active.

Then, the sparsity of coefficients $\bm{\alpha}$ is achieved by incorporating $\ell_1$ regularization into the loss function:
\begin{equation}\label{lasso}
    \rho(\bm{\alpha},\bm{\theta}) = 
    1 - \frac{\bm{Y}^{c,\top}\left(  K_{\alpha,\theta}(\bm{X}^c,\bm{X}^c) +\lambda_1\bm{I}\right)^{-1} \bm{Y}^c}{\bm{Y}^{b,\top} \left( K_{\alpha,\theta}(\bm{X}^b,\bm{X}^b)+\lambda_1\bm{I} \right) ^{-1} \bm{Y}^b} +
    {\lambda}_2\|\bm{\alpha}\|_1
\end{equation}
The proposed approach is distinct from that of the  Sparse
Spectral Kernel Ridge Regression \cite{akian2022learning} which incorporates sparsity into the feature-map representation/construction of the Kernel Flows algorithm.
In this paper, we train the objective function by the SGD optimizer \cite{bottou2012stochastic} implemented in Pytorch. This results in a trade-off between the accurate approximation of the systems and reducing the magnitude of the kernel coefficients, where the trade-off is determined by the regularization penalty $\lambda_2$.

To simultaneously obtain the sparse coefficient of base kernels and intrinsic parameter from data, we solve equation \eqref{lasso} by iterating the following two steps: (a) fix $\bm{\alpha}$, optimize $\bm{\theta}$; then (b), fix $\bm{\theta}$, optimize $\bm{\alpha}$. The procedure is summarized in Algorithm \ref{alg1}.

\begin{algorithm} \label{alg1}
	\begin{spacing}{1.25}
		\caption {Sparse Kernel Flows}
		\SetCommentSty{mycommfont}
		
		\DontPrintSemicolon
		\SetNoFillComment 
		
		\KwIn{time series $(\bm{x_1},\cdots,\bm{x_n})$, embedding delay $\tau$, case kernel $k_{\alpha,\theta}$, hyper-paramter $\lambda_1$, $\lambda_2$}	
		\KwOut{active kernel function}

		Split data into $\left( \bm{X}_1,\cdots,\bm{X}_N \right) $ and  $\left( \bm{Y}_1,\cdots,\bm{Y}_N \right) $, where $N = n-\tau$  \tcp*{prepare data matrix}
		
		Random initialize parameters $\bm{\alpha}, \bm{\theta}$ \tcp*{initial guess of the base kernel}

		\tcc{update estimates by using KF algorithm presented in Section 2.1}
		\For{$\mathrm{epoch} = 1:m$}
		{
			Fix $\bm{\alpha}$, update $\bm{\theta}  \leftarrow \arg\min\limits_{\theta}\rho(\bm{\theta}|\text{Data})$ 
			
			Fix $\bm{\theta}$, update $\bm{\alpha} \leftarrow \arg\min\limits_{\alpha}\rho(\bm{\alpha}|\text{Data})$ 
		} 		
	\end{spacing}
\end{algorithm}

The implementation of the training procedures described above requires three tuning hyper-parameters, namely, the embedding delay $\tau$, the shrinkage parameter $\lambda_1$, and the sparsity penalty $\lambda_2$. The choice of these parameters impacts the accuracy of the training procedure. Here we outline some methods for the hyper-parameter selection.

The shrinkage parameter $\lambda_1$  avoids overfitting in the presence of noisy measurements. The  hyper-parameter $\lambda_2$ determines the strength of the regularization of the model coefficients $\bm{\alpha}$ and thus affects the sparsity of the resulting kernel function. If $\lambda_2$ is too large, the kernel function will be too simple and achieve poor forecasting; if it is too small, the function will be non-sparse and prone to overfitting. Both $\lambda_1$ and  $\lambda_2$ should be chosen using k-fold cross-validation \cite{exterkate2016nonlinear}, which is a natural criterion for out-of-sample forecasting, cf. below for more details. We'll fix $\tau=5$ in our simulations but two methods were proposed in \cite{BHPhysicaD} to choose $\tau$.

\section{Simulations}
In order to probe the generalization ability and the finite sample performance of the Sparse KFs algorithm, we apply our proposal to the dynamical system benchmarks comprising 132 chaotic dynamical systems in \cite{gilpin2021chaos}. Furthermore, the proposed algorithm is compared to the baseline methods.

Benchmarks are computed on the High Performance Computing Platform of Nanjing University of Aeronautics and Astronautics, using Intel Xeon 8358 CPU and 256 GB RAM per node.

\subsection{Data collection and performance criterion}
The time series observations are collected by the precomputed time series on GitHub at \url{https://github.com/williamgilpin/dysts}. We use the default fine granularity setting by \cite{gilpin2021chaos} and 7200 samples are generated for all systems to test the methods.

Given test samples $\left\lbrace \bm{x}_k\right\rbrace^N_{n+1} $ and the predictions $\left\lbrace \hat{\bm{x}}_k\right\rbrace^N_{n+1} $, the forecasting error is measured by two standard time series similarity metrics, symmetric mean absolute percentage error criterion (SMAPE) and Hausdorff Distance (HD) between the true attractor and the reconstructed attractor from data, respectively expressed as
\begin{equation}
	\text{SMAPE}_{\text{test}}[\bm{x}] = \frac{1}{n} \sum_{i=1}^{d}\sum_{k=n+1}^{N} \frac{|\hat{x}^{(i)}_k-x^{(i)}_k|}{\left(|\hat{x}^{(i)}_k|+|x^{(i)}_k| \right)/2 } \times 100\%
\end{equation}
and
\begin{equation}
	d_H\left( \hat{\bm{x}}, \bm{x}\right) =
	\max\left\lbrace 
	\sup\limits_{\hat{x}\in\hat{\bm{x}}} d(\hat{x},\bm{x}),
	\sup\limits_{x\in\bm{x}} d(\hat{\bm{x}},x)
	\right\rbrace 
\end{equation}
where $d(\hat{x},\bm{x}) = \inf\limits_{x\in \bm{x}} d(\hat{x},x)$ quantifies the distance from a point $\hat{x}\in\hat{\bm{x}}$ to the subset $x\in\bm{x}$.

\subsection{Experimental setting}

Considering the computational efficiency and centre of this work, in all 132 dynamical systems, the embedding delay $\tau$ is set to 5, the shrinking parameter $\lambda_1$ is set to 0.05, and the sparsity parameter $\lambda_2$ is determined by the 3-fold cross-validation \cite{jung2018multiple}. In detail, splitting $N$ samples into three part, for given values $\lambda_2 \in \{0, 0.0001, 0.001, 0.01, 0.1, 1, 10\}$, we learn the kernel on the observations of size $\frac{2}{3}N$ that remains when a contiguous block of $\frac{1}{3}N$ samples is removed. The left out samples then are "forecasted" by the learned kernel and compared to the true values. We repeat this procedure 3 times, and then compute the mean of the three SMAPEs of the results to get an average SMAPE for the 3-fold validation. Performing this cross-validation for each candidate value of the parameter set, we select the best one that leads to the smallest SMAPE over the forecasts. Note that $\lambda_2=0$ represents the Regular KFs scenario.

In addition to the hyper-parameters used in the objective function training, this algorithm requires an initial choice of base kernel functions. Here, in all 132 dynamical systems, we used a base kernel function that is a linear combination of 21 kernels:
\begin{equation}
	\begin{aligned}
		k_{\alpha,\theta} = &
		\alpha^2_1\left( \bm{x}^\top\bm{y} + \theta_1^2\right)  + 
		\alpha^2_2 \left( \theta_2^2\bm{x}^\top\bm{y} + \theta_3^2\right) ^{|\theta_4|} 
		+ \alpha^2_3\exp\left( {\frac{-\|x-y\|^2_2}{2\theta^2_5}}\right) + 
		\alpha^2_4\exp\left( {\frac{-\|x-y\|_2}{2\theta^2_6}}\right) \\
		&+ \alpha^2_{5} \exp\left(\frac{-\sin^2\left(\pi\|x-y\|^2_2/\theta_{7}  \right) }{\theta^2_{8}} \right) 
		\exp\left( -\frac{\|x-y\|^2_2}{\theta^2_{9}}\right)
		+ \alpha^2_{6} \exp\left(\frac{-\sin^2\left(\pi\|x-y\|^2_2/\theta_{10}  \right) }{\theta^2_{11}} \right) 
		\\
		& + \alpha^2_{7} \exp\left(\frac{-\sin^2\left(\pi\|x-y\|_2/\theta_{12}  \right) }{\theta^2_{13}} \right) 
		\exp\left( -\frac{\|x-y\|_2}{\theta^2_{14}}\right) 
		+ \alpha^2_{8} \exp\left(\frac{-\sin^2\left(\pi\|x-y\|_2/\theta_{15}  \right) }{\theta^2_{16}} \right) \\
		&   + \alpha^2_9 \left( \|x-y\|^2_2+\theta^2_{17}\right) ^{\frac{1}{2}}
		+ \alpha^2_{10} \left(\theta^2_{18} + \theta^2_{19}\|x-y\|^2_2 \right)^{-\frac{1}{2}}
		+ \alpha^2_{11} \left( \theta^2_{20} + \theta^2_{21}\|x-y\|_2\right)^{-\frac{1}{2}} \\
		&   +\alpha^2_{12}\left(\theta^2_{22}+ \|x-y\|_2\right)^{\theta_{23}}
		+ \alpha^2_{13} \left( \theta^2_{24} + \|x-y\|_2^2\right)^{\theta_{25}} 
		+ \alpha^2_{14} \left(1+\left( \frac{\|x-y\|_2}{\theta_{26}}\right) ^2 \right)^{-1}\\
		& + \alpha^2_{15} \left( 1+ \frac{\|x-y\|_2}{\theta^2_{27}} \right)^{-1} 
		+ \alpha^2_{16}\left(1- \frac{\|x-y\|^2_2}{\|x-y\|^2_2+\theta^2_{28}} \right)\\
		&	+ \alpha^2_{17} \max\left( {0,1-\frac{\|x-y\|^2_2}{\theta^2_{29}}}\right)
		+ \alpha^2_{18} \max\left( {0,1-\frac{\|x-y\|_2}{\theta^2_{30}}}\right) 
		\\
		&	
		+\alpha^2_{19}\left(\log(\|x-y\|^{\theta_{31}}_2+1)\right)
		+ \alpha^2_{20} \tanh\left( \theta_{32}\bm{x}^\top\bm{y} + \theta_{33}\right) \\
		&+ \alpha^2_{21} \left[ \arccos\left( -\frac{\|x-y\|_2}{\theta^2_{34}}\right)-  \frac{\|x-y\|_2}{\theta^2_{34}}\sqrt{1-\left( \frac{\|x-y\|_2}{\theta^2_{34}}\right)^2 }\right] \cdot \bm{1}_{\left\{x,y: ||x-y||_2^2 < \theta^2_{34}  \right\} }
	\end{aligned}
\end{equation}

\subsection{Examples}
\begin{figure}[!ht]
	\centering
	\includegraphics[width=.97\linewidth]{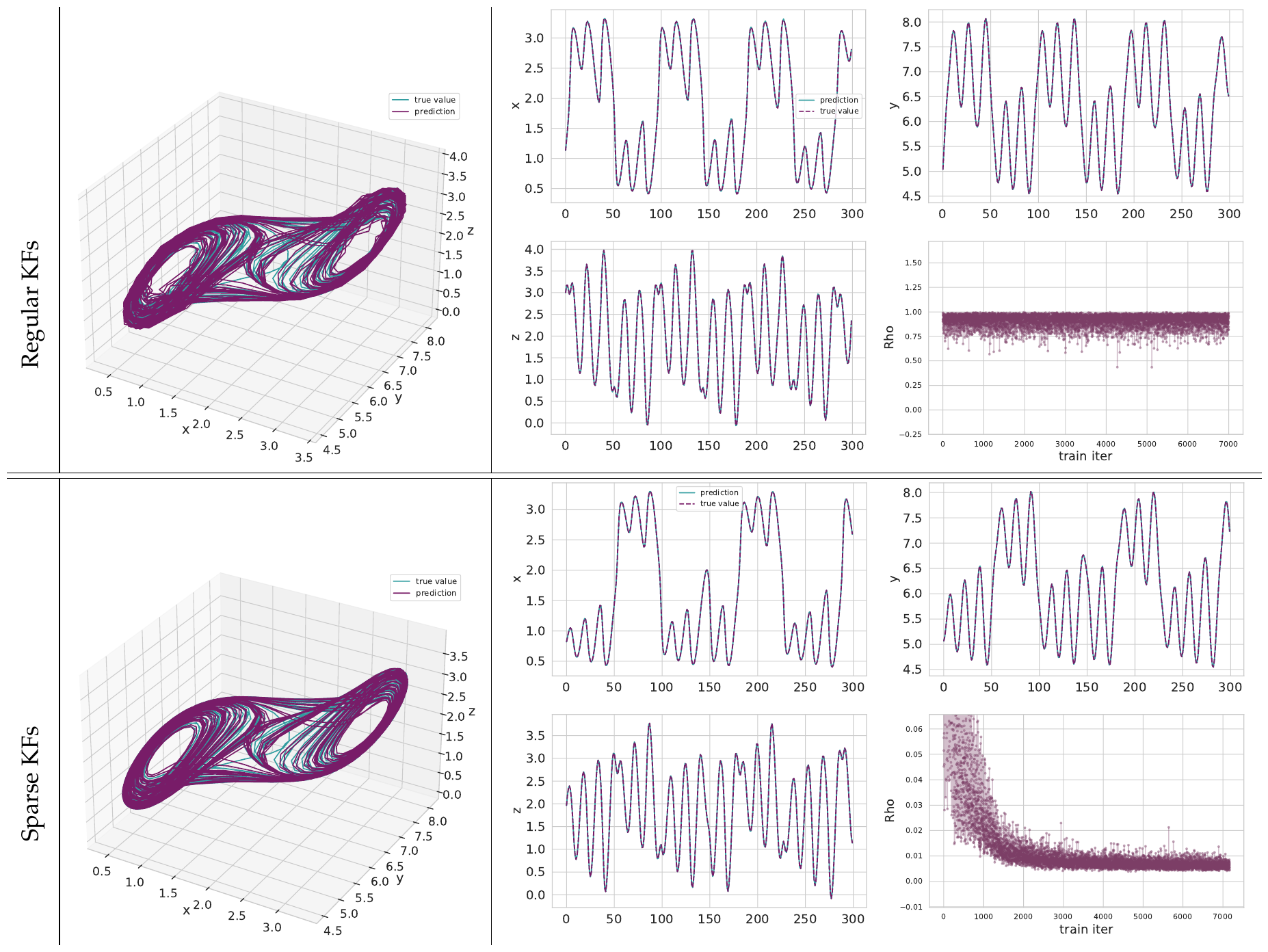}
	\caption{A comparison of learned phase, trajectories, and loss for the BeerRNN system.}
	\label{Beer}
\end{figure}

\begin{figure}[!ht]
	\centering
	\includegraphics[width=.97\linewidth]{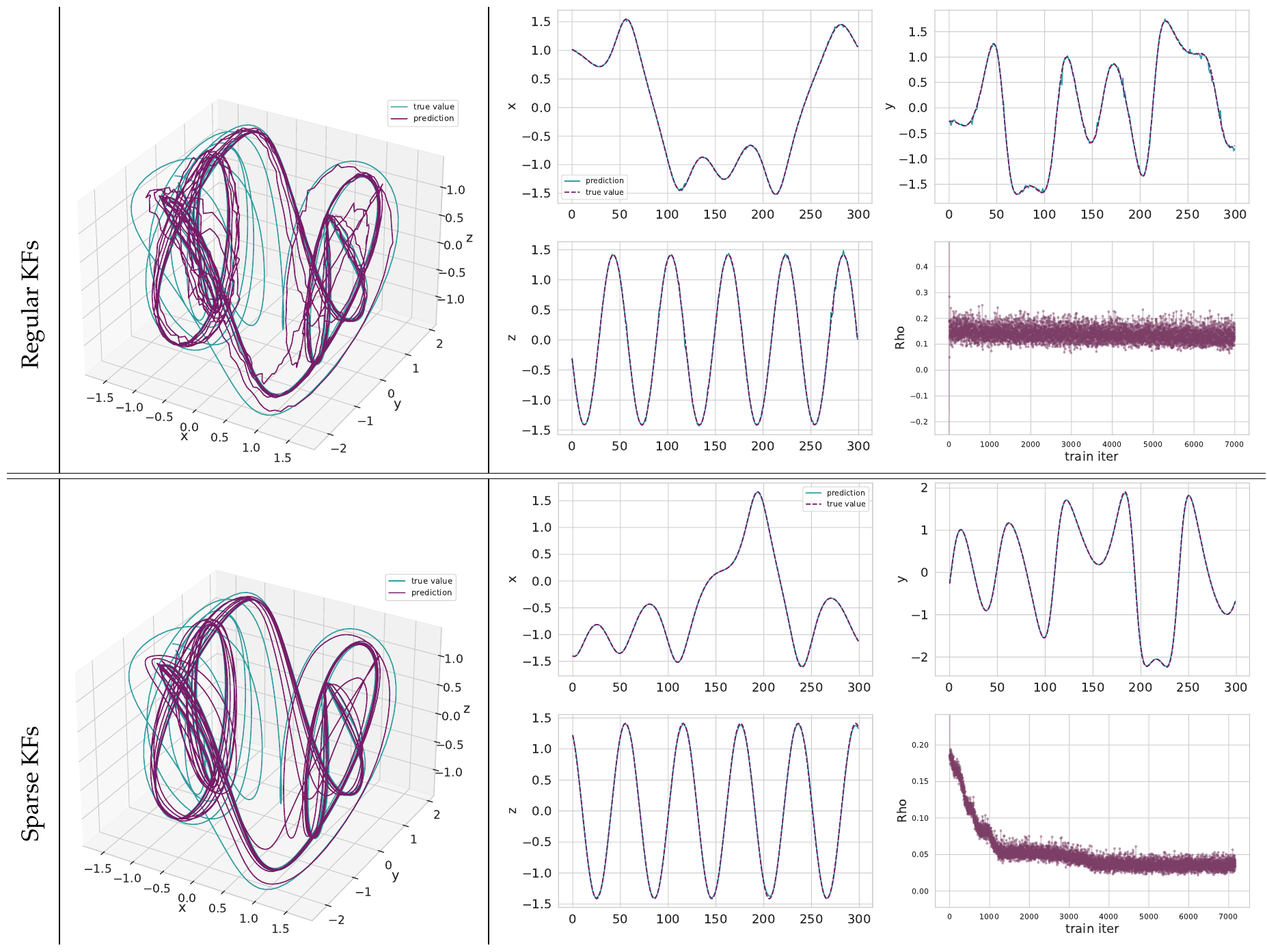}
	\caption{A comparison of learned phase, trajectories, and loss for the Duffing system.}
	\label{Duffing}
\end{figure}

\begin{figure}[!ht]
	\centering
	\includegraphics[width=.97\linewidth]{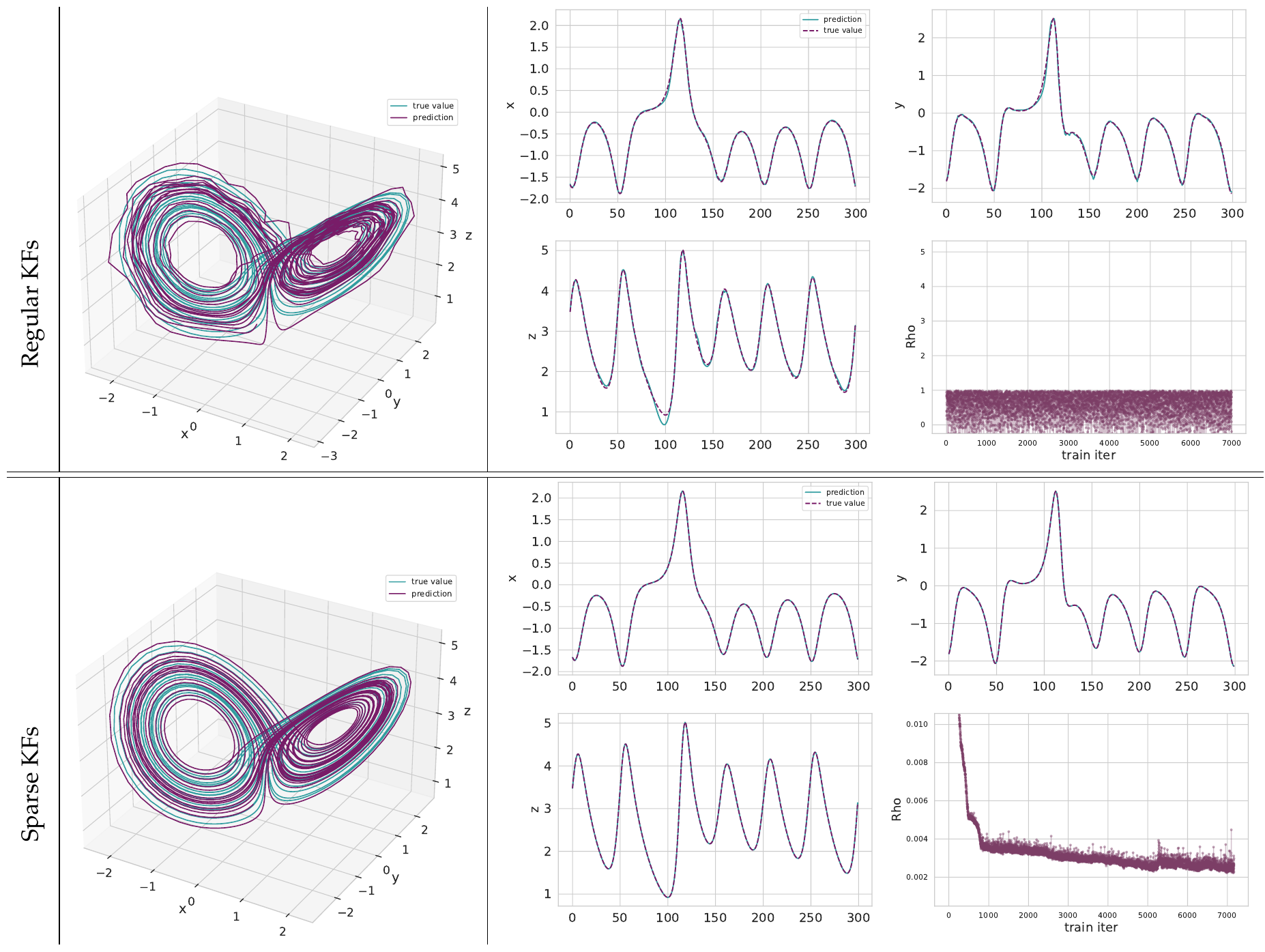}
	\caption{A comparison of learned phase, trajectories, and loss for the Lorenz system.}
	\label{Lorenz}
\end{figure}

In order to demonstrate the learning results intuitively, we present three examples here, including BeerRNN, Duffing,  and Lorenz systems. The identification results obtained from  Regular and Sparse KFs are summarized in Figures \ref{Beer}-\ref{Lorenz}, and Tables \ref{erro}-\ref{coeff}.

Figures \ref{Beer}-\ref{Lorenz} present the learned phase plot, trajectories, and loss iterations of three dynamics. It is clear that in all cases, the approximated phases and trajectories by Sparse Kfs almost coincide with the true ones although the test systems are chaotic. This phenomenon partially indicates the superiority of this kind of learning method for the reconstruction of discrete maps. However, there exist some outliers of learned trajectories obtained by Regular KFs. 
Besides, the value loss of Sparse KFs shows a more pronounced downward trend than that of Regular KFs as the training step increases, indicating the superiority of Sparse KFs in terms of convergence. 

Table \ref{erro} shows the forecasting error. For all systems, Sparse KFs has smaller SMAPEs and HDs, indicating better forecasting performance. In addition, Table \ref{coeff} shows that the weight coefficient obtained by the Regular Kfs are non-zeros, whereas the Sparse version gets sparsity weight coefficients, indicating the $L_1$ regularization penalty on $\bm{\alpha}$ promotes sparsity in the resulting learned kernel function.

\begin{table}[H]
	\caption{Forecasting errors obtained by the Regular KFs and Sparse KFs}
	\label{erro}
	\setlength{\tabcolsep}{8mm}
	\begin{tabular}{|l||c|c||c|c|}
		\hline
		name    & \multicolumn{2}{|c||}{Regular KFs}                    & \multicolumn{2}{c|}{Sparse KFs}                     \\
		\cline{2-5} 
		& \multicolumn{1}{c|}{SMAPE} & \multicolumn{1}{c||}{HD} & \multicolumn{1}{c|}{SMAPE} & \multicolumn{1}{c|}{HD} \\
		\hline
		BeerRNN & 0.0284                    & 0.1164                 & 0.0048                    & 0.0333                 \\
		\hline
		Duffing & 0.0560                    & 0.0425                 & 0.0075                    & 0.0043                 \\
		\hline
		Lorenz  & 0.0372                    & 0.0130                 & 0.0097                    & 0.0056            \\
		\hline    
	\end{tabular}
\end{table}

\begin{table}[H]
	\footnotesize  
	\caption{Weight coefficients of kernel obtained by the Regular KFs and Sparse KFs}
	\label{coeff}
	\setlength{\tabcolsep}{6mm}
	\begin{tabular}{crrrrrr}
		\toprule
		Coefficient & \multicolumn{2}{c}{BeerRNN} & \multicolumn{2}{c}{Duffing} & \multicolumn{2}{c}{Lorenz}  \\
		\cmidrule(lr){2-3} \cmidrule(lr){4-5} \cmidrule(lr){6-7} 
		& Regular   & Sparse   & Regular  & Sparse  & Regular   & Sparse  \\
		\midrule
		$\alpha_1$  & 0.5746 & 0      & 0.7285 & 0      & 0.7341 & 0     \\
		$\alpha_2$  & 0.4692 & 0.9985 & 0.7310 & 0.0242 & 0.7284 & 0.9918     \\
		$\alpha_3$  & 0.7234 & 0      & 0.7170 & 0.6807 & 0.7341 & 0           \\
		$\alpha_4$  & 0.7316 & 0      & 0.7238 & 0      & 0.7355 & 0         \\
		$\alpha_5$  & 0.7474 & 0      & 0.7008 & 0      & 0.7298 & 0         \\
		$\alpha_6$  & 0.7180 & 0      & 0.8214 & 0      & 0.6484 & 0       \\
		$\alpha_7$  & 0.7448 & 0      & 0.7234 & 0      & 0.7442 & 0.2498     \\
		$\alpha_8$  & 0.8351 & 0      & 0.7272 & 0      & 0.7206 & 0           \\
		$\alpha_9$  & 0.8604 & 0.9998 & 0.7658 & 0      & 0.7450 & 0        \\
		$\alpha_{10}$ & 0.7660 & 0      & 0.7239 & 0      & 0.7362 & 0          \\
		$\alpha_{11}$ & 0.7343 & 0      & 0.7265 & 0      & 0.7376 & 0          \\
		$\alpha_{12}$ & 0.7527 & 0      & 0.7310 & 0      & 0.7346 & 0.2785     \\
		$\alpha_{13}$ & 0.7337 & 0      & 0.7063 & 0      & 0.7391 & 0        \\
		$\alpha_{14}$ & 0.7322 & 0      & 0.7116 & 0      & 0.7351 & 0        \\
		$\alpha_{15}$ & 0.7398 & 0      & 0.7027 & 0      & 0.7433 & 0       \\
		$\alpha_{16}$ & 0.7363 & 0      & 0.7031 & 0      & 0.7333 & 0          \\
		$\alpha_{17}$ & 0.7052 & 0      & 0.7131 & 0      & 0.7506 & 0          \\
		$\alpha_{18}$ & 0.6815 & 0      & 0.6978 & 0      & 0.7497 & 0          \\
		$\alpha_{19}$ & 0.6959 & 0.8630 & 0.7676 & 0      & 0.7343 & 0        \\
		$\alpha_{20}$ & 0.7305 & 1      & 0.7789 & 0      & 0.6839 & 0        \\
		$\alpha_{21}$ & 0.7321 & 0      & 0.7303 & 0      & 0.7310 & 0         \\
		\bottomrule
	\end{tabular}
\end{table}

\newpage
\subsection{Comparison with baseline methods}
\begin{figure}[!ht]
	\centering
	\includegraphics[width=1\linewidth]{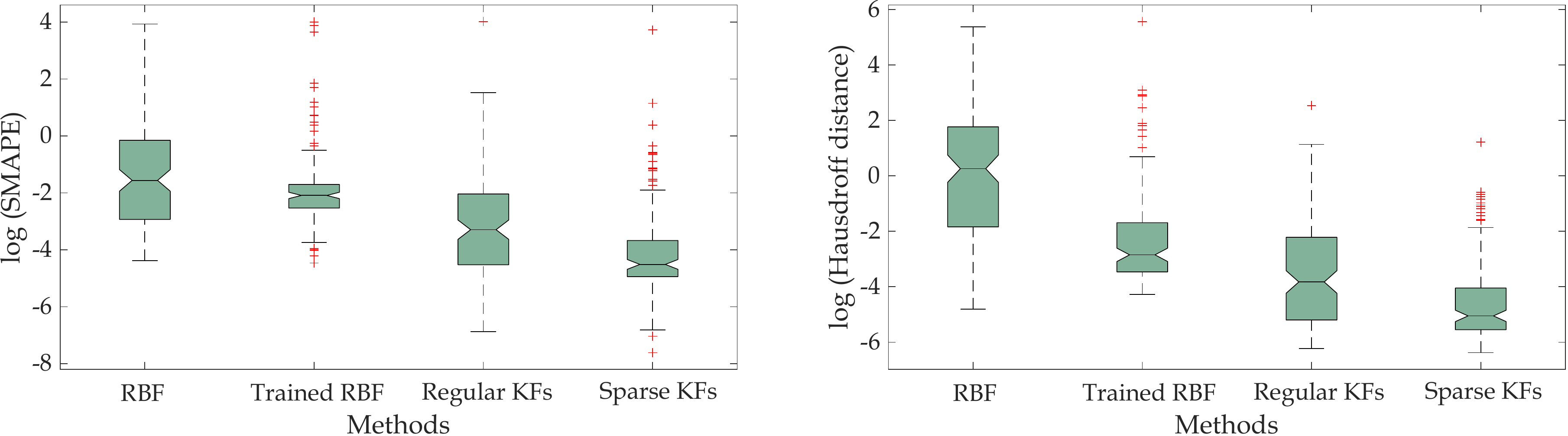}
	\caption{Distribution of forecasting errors for different methods for all 132 dynamical systems.}
	\label{methods}
\end{figure}

In order to further demonstrate the performance of the method, we make a comparison with several baseline methods, including kernel regression with Gaussian kernel ($\sigma=0.5$) and trained Gaussian kernel by regular KFs. For the convenience of presentation, the two methods are abbreviated to ``RBF'', and ``Trained RBF'', respectively. The  comparison results are summarized at Figure \ref{methods} and Table 2.

On the basis of the results, we may roughly conclude the following: Figure \ref{methods} presents the forecasting performance of four methods, and every box depicts the distribution of forecasting errors for all 132 dynamical systems. From the statistical perspective, the performance of the four methods can be summarized as
\[
\text{RBF} \prec \text{Trained RBF} \prec \text{Regular KFs} \prec \text{RBF} 
\] 
Specifically, Table 2 shows the detailed forecasting errors. In terms of the SMAPE criterion, in 120 dynamical systems, Sparse KFs has the smallest SMPEs, indicating adaptive kernel provides more precious forecasts. The Trained RBF performs best in 9 systems, partially indicating the kernel learning methods could improve the forecasting performance of kernel regression. In addition, there are 3 systems which Regular Kfs performs best with smallest SMAPEs.  

\newpage
\begin{center} 
	\tablefirsthead{%
		\multicolumn{8}{l}{\textbf{Table 2: one-head forecasting errors for 132 chaotic dynamical systems}}\\
		\toprule		
		Name                    & \multicolumn{2}{l}{RBF Kernel}  & \multicolumn{2}{l}{Trained RBF Kernel} & \multicolumn{2}{l}{Regular KFs} & \multicolumn{2}{l}{Sparse KFs} & Best         \\
		\cmidrule(lr){2-3} \cmidrule(lr){4-5} \cmidrule(lr){6-7} \cmidrule(lr){8-9}
		& SMAPE & HD  	& SMAPE & HD    	& SMAPE & HD		& SMAPE & HD  & \\
		\midrule }
	\tablehead{%
		\hline
		\multicolumn{10}{l}{\small\sl continued from previous page}\\
		\hline
		Name                  &  \multicolumn{2}{l}{RBF Kernel}  & \multicolumn{2}{l}{Trained RBF Kernel} & \multicolumn{2}{l}{Regular KFs} & \multicolumn{3}{l}{Sparse KFs}        \\
		\cmidrule(lr){2-3} \cmidrule(lr){4-5} \cmidrule(lr){6-7} \cmidrule(lr){8-10}
		& SMAPE & HD  	& SMAPE & HD    	& SMAPE & HD		& SMAPE & HD  & Best \\
		\midrule}
	\tabletail{%
		\hline
		\multicolumn{10}{r}{\small\sl continued on next page}\\
		\hline}
	\tablelasttail{ \bottomrule }
	
	\footnotesize
	\begin{supertabular}{llllllllll}
		
		Aizawa                  & 0.2186  & 1.2855   & 0.2049  & 0.0244   & 0.0683  & 0.0081  & 0.0322  & 0.0027 & Sparse KFs  \\
		AnishchenkoAstakhov     & 0.1554  & 2.2373   & 0.0927  & 0.0507   & 0.0312  & 0.0190  & 0.0082  & 0.0030 & Sparse KFs  \\
		Arneodo                 & 0.0795  & 0.0472   & 0.1028  & 0.0239   & 0.0106  & 0.0086  & 0.0103  & 0.0027 & Sparse KFs  \\
		ArnoldBeltramiChildress & 0.0456  & 0.0478   & 0.1443  & 0.0297   & 0.0832  & 0.0179  & 0.0764  & 0.0173 & Trained RBF \\
		ArnoldWeb               & 24.2023 & 171.4064 & 48.4955 & 262.1925 & 0.0391  & 0.1522  & 0.0160  & 0.2081 & Sparse KFs  \\
		AtmosphericRegime       & 0.4452  & 0.6260   & 0.6062  & 0.4058   & 0.5512  & 0.7474  & 0.5168  & 0.3314 & Sparse KFs  \\
		BeerRNN                 & 0.3792  & 10.1832  & 0.1477  & 0.2636   & 0.0284  & 0.1164  & 0.0048  & 0.0333 & Sparse KFs  \\
		BelousovZhabotinsky     & 0.1022  & 215.8174 & 0.1203  & 6.1369   & 0.1610  & 0.1173  & 0.0096  & 0.0304 & Sparse KFs  \\
		BickleyJet              & 0.0159  & 0.0770   & 0.0833  & 0.0620   & 0.0384  & 0.0311  & 0.0043  & 0.0065 & Sparse KFs  \\
		Blasius                 & 0.0627  & 1.0313   & 0.0242  & 0.0468   & 0.0092  & 0.0089  & 0.0083  & 0.0061 & Sparse KFs  \\
		BlinkingRotlet          & 0.7304  & 0.9160   & 0.7737  & 0.4399   & 0.7327  & 0.8686  & 0.2059  & 0.4307 & Trained RBF \\
		BlinkingVortex          & 1.4024  & 0.8494   & 1.6182  & 0.5333   & 0.7504  & 0.5914  & 0.5423  & 0.5566 & Trained RBF \\
		Bouali                  & 0.2115  & 4.5773   & 0.0794  & 0.0584   & 0.1644  & 0.1649  & 0.0057  & 0.0075 & Sparse KFs  \\
		Bouali2                 & 0.1604  & 3.6113   & 0.0468  & 0.0411   & 0.0010  & 0.0355  & 0.0034  & 0.0086 & Sparse KFs  \\
		BurkeShaw               & 1.0260  & 5.2350   & 0.0879  & 0.0648   & 0.0123  & 0.0052  & 0.0099  & 0.0034 & Sparse KFs  \\
		CaTwoPlus               & 0.0188  & 0.3077   & 0.0148  & 0.0908   & 0.0025  & 0.0220  & 0.0011  & 0.0119 & Sparse KFs  \\
		CaTwoPlusQuasiperiodic  & 0.0263  & 0.1421   & 0.0240  & 0.1012   & 0.0022  & 0.0175  & 0.0005  & 0.0049 & Sparse KFs  \\
		CellCycle               & 1.3828  & 15.5649  & 0.0236  & 0.1439   & 0.0144  & 0.1163  & 0.0049  & 0.0228 & Sparse KFs  \\
		CellularNeuralNetwork   & 0.0499  & 0.0132   & 0.1462  & 0.0307   & 0.8356  & 0.2267  & 0.0326  & 0.0088 & RBF         \\
		Chen                    & 0.7046  & 12.0451  & 0.0736  & 0.0776   & 0.0100  & 0.0073  & 0.0125  & 0.0066 & Sparse KFs  \\
		ChenLee                 & 0.0813  & 0.1641   & 0.1314  & 0.0568   & 0.3262  & 0.0269  & 0.0075  & 0.0036 & Sparse KFs  \\
		Chua                    & 0.0388  & 0.0122   & 0.1237  & 0.0287   & 0.0188  & 0.0068  & 0.0031  & 0.0064 & RBF         \\
		CircadianRhythm         & 0.2496  & 2.5516   & 0.2166  & 0.0587   & 0.0725  & 0.0735  & 0.1488  & 0.0418 & Sparse KFs  \\
		CoevolvingPredatorPrey  & 0.2718  & 4.5867   & 0.3627  & 0.0489   & 0.1599  & 0.1538  & 0.0190  & 0.0460 & Trained RBF \\
		Colpitts                & 0.0394  & 0.9218   & 0.0608  & 0.1672   & 0.0544  & 0.1106  & 0.0019  & 0.0339 & Sparse KFs  \\
		Coullet                 & 0.0666  & 0.2871   & 0.0943  & 0.0313   & 0.0155  & 0.0095  & 0.0031  & 0.0027 & Regular KFs \\
		Dadras                  & 1.9320  & 9.5681   & 0.3464  & 2.7592   & 0.1888  & 0.0612  & 0.0292  & 0.0218 & Sparse KFs  \\
		DequanLi                & 0.2080  & 1.0061   & 0.1323  & 0.0296   & 0.0102  & 0.0040  & 0.0076  & 0.0034 & Sparse KFs  \\
		DoubleGyre              & 0.1868  & 0.4850   & 0.0945  & 0.0364   & 0.1599  & 0.1428  & 0.0054  & 0.0077 & Sparse KFs  \\
		DoublePendulum          & 7.5702  & 8.2609   & 0.7037  & 0.4982   & 1.0474  & 0.5119  & 0.7049  & 0.4731 & Sparse KFs  \\
		Duffing                 & 0.3260  & 0.8533   & 0.1883  & 0.0375   & 0.0560  & 0.0425  & 0.0075  & 0.0043 & Sparse KFs  \\
		ExcitableCell           & 0.3806  & 83.0296  & 0.0582  & 22.1067  & 0.0115  & 0.1725  & 0.0011  & 0.0125 & Sparse KFs  \\
		Finance                 & 0.0980  & 1.9899   & 0.0523  & 0.0912   & 0.0823  & 0.0557  & 0.0095  & 0.0062 & Sparse KFs  \\
		FluidTrampoline         & 0.1125  & 0.0568   & 0.2337  & 0.0454   & 0.0140  & 0.0189  & 0.0076  & 0.0114 & Sparse KFs  \\
		ForcedBrusselator       & 0.0774  & 1.3912   & 0.0417  & 0.0644   & 0.0293  & 0.0528  & 0.0036  & 0.0109 & Sparse KFs  \\
		ForcedFitzHughNagumo    & 0.0765  & 0.1537   & 0.1406  & 0.0251   & 0.0980  & 0.1159  & 0.0595  & 0.0363 & Trained RBF \\
		ForcedVanDerPol         & 0.1977  & 0.9019   & 0.1815  & 0.4087   & 0.7877  & 0.4033  & 0.1758  & 0.0333 & Sparse KFs  \\
		GenesioTesi             & 0.0303  & 0.0250   & 0.0790  & 0.0229   & 0.0038  & 0.0034  & 0.0099  & 0.0035 & Sparse KFs  \\
		GuckenheimerHolmes      & 0.3252  & 1.8615   & 0.1818  & 0.0189   & 0.0163  & 0.0045  & 0.0113  & 0.0021 & Sparse KFs  \\
		Hadley                  & 0.1454  & 1.2400   & 0.0958  & 0.0261   & 0.1057  & 0.0303  & 0.0070  & 0.0027 & Sparse KFs  \\
		Halvorsen               & 0.0476  & 0.0394   & 0.0850  & 0.0278   & 0.8315  & 0.2636  & 0.0163  & 0.0036 & Sparse KFs  \\
		HastingsPowell          & 0.0156  & 0.1860   & 0.0183  & 0.0814   & 0.0070  & 0.0234  & 0.0039  & 0.0093 & Sparse KFs  \\
		HenonHeiles             & 0.0740  & 0.0255   & 0.1751  & 0.0339   & 0.0130  & 0.0020  & 0.0262  & 0.0020 & Sparse KFs  \\
		HindmarshRose           & 0.1705  & 2.2443   & 0.1567  & 0.0941   & 0.1879  & 0.0676  & 0.0501  & 0.0121 & Sparse KFs  \\
		Hopfield                & 0.2690  & 0.5214   & 0.2258  & 0.0654   & 0.0704  & 0.0859  & 0.0683  & 0.0816 & Trained RBF \\
		HyperBao                & 0.8635  & 8.3261   & 0.1094  & 0.0695   & 0.0971  & 0.0169  & 0.0020  & 0.0062 & Sparse KFs  \\
		HyperCai                & 0.2774  & 3.7645   & 0.1040  & 0.0698   & 0.1104  & 0.0939  & 0.0073  & 0.0072 & Sparse KFs  \\
		HyperJha                & 0.5032  & 5.8067   & 0.1576  & 0.0658   & 0.0825  & 0.0365  & 0.0091  & 0.0055 & Sparse KFs  \\
		HyperLorenz             & 1.9790  & 12.2844  & 0.1806  & 0.0792   & 0.0166  & 0.0064  & 0.0216  & 0.0060 & Sparse KFs  \\
		HyperLu                 & 1.2429  & 10.7042  & 0.1245  & 0.0689   & 0.1049  & 0.0732  & 0.0074  & 0.0061 & Sparse KFs  \\
		HyperPang               & 0.7841  & 7.2542   & 0.1633  & 0.0824   & 0.0473  & 0.0208  & 0.0182  & 0.0070 & Sparse KFs  \\
		HyperQi                 & 4.5864  & 19.6760  & 2.7646  & 17.6765  & 0.0092  & 0.0052  & 0.0096  & 0.0042 & Sparse KFs  \\
		HyperRossler            & 0.8076  & 12.6803  & 0.1299  & 0.3022   & 0.2940  & 0.0146  & 0.0321  & 0.0066 & Sparse KFs  \\
		HyperWang               & 1.7118  & 14.8686  & 0.1473  & 0.0338   & 0.0027  & 0.0026  & 0.0069  & 0.0047 & Sparse KFs  \\
		HyperXu                 & 2.2143  & 19.1536  & 0.1773  & 0.2483   & 0.0090  & 0.0120  & 0.0197  & 0.0077 & Sparse KFs  \\
		HyperYan                & 1.0246  & 12.2693  & 0.2782  & 5.2669   & 0.0563  & 0.0265  & 0.0104  & 0.0037 & Sparse KFs  \\
		HyperYangChen           & 1.5399  & 16.1034  & 0.1534  & 0.1838   & 0.0113  & 0.0068  & 0.0247  & 0.0066 & Sparse KFs  \\
		IkedaDelay              & 4.6646  & 5.4753   & 6.3582  & 6.6615   & 1.1826  & 0.1482  & 0.2961  & 0.0856 & Sparse KFs  \\
		IsothermalChemical      & 0.0235  & 1.3498   & 0.0192  & 0.1858   & 0.0044  & 0.0315  & 0.0018  & 0.0127 & Sparse KFs  \\
		ItikBanksTumor          & 3.1791  & 1.5271   & 5.4786  & 0.0394   & 1.9552  & 0.0032  & 1.4649  & 0.0033 & Sparse KFs  \\
		JerkCircuit             & 0.2332  & 1.4263   & 0.3403  & 1.0682   & 0.1393  & 0.4640  & 0.0815  & 0.2651 & Sparse KFs  \\
		KawczynskiStrizhak      & 0.0176  & 0.0126   & 0.0405  & 0.0150   & 0.0330  & 0.0459  & 0.0029  & 0.0062 & RBF         \\
		Laser                   & 0.0920  & 0.1182   & 0.1116  & 0.0297   & 0.0056  & 0.0033  & 0.0046  & 0.0039 & Sparse KFs  \\
		LiuChen                 & 50.8445 & 13.2288  & 54.7053 & 11.5904  & 55.5113 & 3.1109  & 41.2576 & 0.5160 & Sparse KFs  \\
		Lorenz                  & 0.0441  & 0.4909   & 0.0684  & 0.0484   & 0.0372  & 0.0130  & 0.0097  & 0.0056 & Sparse KFs  \\
		Lorenz84                & 0.2626  & 0.2467   & 0.1896  & 0.0322   & 0.1597  & 0.0992  & 0.0198  & 0.0047 & Sparse KFs  \\
		Lorenz96                & 1.5382  & 1.7985   & 0.3046  & 0.0576   & 0.3691  & 0.1072  & 0.0682  & 0.0069 & Sparse KFs  \\
		LorenzBounded           & 0.0685  & 0.4453   & 0.0597  & 0.0467   & 0.0217  & 0.0875  & 0.0108  & 0.0042 & Sparse KFs  \\
		LorenzCoupled           & 2.0416  & 12.6271  & 0.1269  & 0.1778   & 0.0017  & 0.0054  & 0.0106  & 0.0077 & Sparse KFs  \\
		LorenzStenflo           & 0.4044  & 7.6345   & 0.1226  & 0.0887   & 0.0166  & 0.0214  & 0.0236  & 0.0075 & Sparse KFs  \\
		LuChen                  & 0.0531  & 1.3090   & 0.0527  & 0.0665   & 0.0717  & 0.0318  & 0.0112  & 0.0068 & Sparse KFs  \\
		LuChenCheng             & 0.0295  & 0.0643   & 0.0991  & 0.0342   & 0.0126  & 0.0161  & 0.0048  & 0.0036 & Sparse KFs  \\
		MacArthur               & 4.2470  & 39.7701  & 0.1105  & 0.7623   & 0.0471  & 0.5928  & 0.0120  & 0.2382 & Sparse KFs  \\
		MackeyGlass             & 3.4885  & 16.1632  & 0.0543  & 0.2721   & 0.0083  & 0.0722  & 0.0211  & 0.0713 & Sparse KFs  \\
		MooreSpiegel            & 0.0734  & 0.1170   & 0.1194  & 0.0340   & 0.0418  & 0.0195  & 0.0120  & 0.0044 & Sparse KFs  \\
		MultiChua               & 0.0255  & 0.0235   & 0.0662  & 0.0341   & 0.0050  & 0.0068  & 0.0074  & 0.0079 & Sparse KFs  \\
		NewtonLiepnik           & 0.2729  & 1.5179   & 0.0777  & 0.0482   & 0.0279  & 0.0127  & 0.0080  & 0.0064 & Sparse KFs  \\
		NoseHoover              & 0.2481  & 1.2736   & 0.1457  & 0.0472   & 0.0055  & 0.0023  & 0.0168  & 0.0031 & Sparse KFs  \\
		NuclearQuadrupole       & 0.8340  & 1.3394   & 0.1462  & 0.0281   & 0.0307  & 0.0132  & 0.0087  & 0.0035 & Sparse KFs  \\
		OscillatingFlow         & 0.7238  & 1.3119   & 0.2332  & 0.0722   & 0.1812  & 0.0769  & 0.1080  & 0.0362 & Sparse KFs  \\
		PanXuZhou               & 0.0851  & 0.2774   & 0.1313  & 0.0493   & 0.0379  & 0.0048  & 0.0104  & 0.0046 & Sparse KFs  \\
		PehlivanWei             & 0.2546  & 1.1650   & 0.1160  & 0.0385   & 0.0183  & 0.0043  & 0.0122  & 0.0027 & Sparse KFs  \\
		PiecewiseCircuit        & 11.9312 & 9.5461   & 1.1743  & 1.2830   & 0.4631  & 0.3629  & 0.4094  & 0.3498 & Sparse KFs  \\
		Qi                      & 2.3154  & 49.6780  & 0.1436  & 1.5068   & 1.0231  & 0.3504  & 0.0414  & 0.0185 & Sparse KFs  \\
		QiChen                  & 0.1621  & 6.9077   & 0.0384  & 0.0619   & 0.1220  & 0.0412  & 0.0137  & 0.0099 & Sparse KFs  \\
		RabinovichFabrikant     & 0.0308  & 0.0514   & 0.0403  & 0.0420   & 0.0058  & 0.0045  & 0.0061  & 0.0046 & Sparse KFs  \\
		RayleighBenard          & 0.1740  & 4.0007   & 0.0796  & 0.0566   & 0.0086  & 0.0065  & 0.0087  & 0.0067 & Sparse KFs  \\
		RikitakeDynamo          & 0.4957  & 5.5948   & 0.0868  & 0.0770   & 0.0686  & 0.0367  & 0.0101  & 0.0042 & Sparse KFs  \\
		Rossler                 & 0.3387  & 4.7017   & 0.1132  & 0.1881   & 0.0318  & 0.0034  & 0.0295  & 0.0045 & Sparse KFs  \\
		Rucklidge               & 0.2231  & 2.8533   & 0.0956  & 0.0824   & 0.0125  & 0.0055  & 0.0142  & 0.0053 & Sparse KFs  \\
		Sakarya                 & 1.1720  & 7.4699   & 0.1405  & 0.4412   & 0.0126  & 0.0046  & 0.0154  & 0.0050 & Sparse KFs  \\
		SaltonSea               & 0.0149  & 0.0643   & 0.0180  & 0.0514   & 0.0038  & 0.0086  & 0.0009  & 0.0039 & Sparse KFs  \\
		SanUmSrisuchinwong      & 0.3718  & 2.4040   & 0.2000  & 0.0462   & 0.0393  & 0.0172  & 0.0172  & 0.0086 & Sparse KFs  \\
		ScrollDelay             & 1.6963  & 2.1868   & 0.3892  & 0.0714   & 1.4291  & 0.1713  & 0.3249  & 0.1553 & Trained RBF \\
		ShimizuMorioka          & 0.0384  & 0.0904   & 0.0977  & 0.0408   & 0.0077  & 0.0051  & 0.0128  & 0.0048 & Sparse KFs  \\
		SprottA                 & 0.1883  & 0.7459   & 0.1464  & 0.0227   & 0.0107  & 0.0048  & 0.0150  & 0.0033 & Sparse KFs  \\
		SprottB                 & 1.5516  & 7.6672   & 0.1158  & 0.3188   & 0.1499  & 0.0434  & 0.0108  & 0.0039 & Sparse KFs  \\
		SprottC                 & 0.0647  & 0.4746   & 0.0962  & 0.0276   & 0.0080  & 0.0038  & 0.0074  & 0.0032 & Sparse KFs  \\
		SprottD                 & 0.2941  & 3.3511   & 0.0625  & 0.0472   & 0.0110  & 0.0049  & 0.0071  & 0.0048 & Sparse KFs  \\
		SprottDelay             & 0.0332  & 0.2678   & 0.0115  & 0.0158   & 0.0565  & 0.3096  & 0.0018  & 0.0050 & Sparse KFs  \\
		SprottE                 & 0.0749  & 0.7674   & 0.0916  & 0.0312   & 0.0023  & 0.0024  & 0.0124  & 0.0042 & Sparse KFs  \\
		SprottF                 & 0.0882  & 0.6506   & 0.1148  & 0.0237   & 0.0079  & 0.0047  & 0.0083  & 0.0046 & Regular KFs \\
		SprottG                 & 0.0427  & 0.0242   & 0.1096  & 0.0287   & 0.0242  & 0.0054  & 0.0155  & 0.0034 & Sparse KFs  \\
		SprottH                 & 0.2228  & 0.9717   & 0.1818  & 0.0444   & 0.0139  & 0.0055  & 0.0178  & 0.0051 & Sparse KFs  \\
		SprottI                 & 0.0302  & 0.0316   & 0.0672  & 0.0241   & 0.0517  & 0.0490  & 0.0055  & 0.0038 & Sparse KFs  \\
		SprottJ                 & 0.0175  & 0.0141   & 0.0685  & 0.0189   & 0.1167  & 0.0140  & 0.0090  & 0.0031 & Sparse KFs  \\
		SprottJerk              & 0.0535  & 0.3400   & 0.1402  & 0.0167   & 0.0449  & 0.0042  & 0.0149  & 0.0042 & Sparse KFs  \\
		SprottK                 & 0.0884  & 0.2206   & 0.1267  & 0.0244   & 0.0168  & 0.0034  & 0.0095  & 0.0030 & Sparse KFs  \\
		SprottL                 & 0.0314  & 0.0744   & 0.0441  & 0.0474   & 0.0041  & 0.0138  & 0.0059  & 0.0083 & Sparse KFs  \\
		SprottM                 & 0.0468  & 0.0187   & 0.0714  & 0.0266   & 0.0101  & 0.0036  & 0.0094  & 0.0034 & Sparse KFs  \\
		SprottMore              & 1.3180  & 1.7383   & 2.0323  & 0.5184   & 0.6949  & 0.3744  & 0.3135  & 0.3061 & Sparse KFs  \\
		SprottN                 & 0.0266  & 0.0081   & 0.1237  & 0.0138   & 0.0870  & 0.0216  & 0.0201  & 0.0035 & Sparse KFs  \\
		SprottO                 & 0.0373  & 0.0191   & 0.1060  & 0.0274   & 0.0169  & 0.0049  & 0.0120  & 0.0024 & Sparse KFs  \\
		SprottP                 & 0.0471  & 0.1090   & 0.1495  & 0.0227   & 0.0096  & 0.0055  & 0.0065  & 0.0018 & Sparse KFs  \\
		SprottQ                 & 0.0541  & 0.0408   & 0.0962  & 0.0234   & 0.0142  & 0.0034  & 0.0037  & 0.0017 & Sparse KFs  \\
		SprottR                 & 0.1182  & 1.2301   & 0.1157  & 0.0314   & 0.0390  & 0.0303  & 0.0122  & 0.0046 & Sparse KFs  \\
		SprottS                 & 0.0436  & 0.0215   & 0.1422  & 0.0283   & 0.0121  & 0.0049  & 0.0072  & 0.0033 & Sparse KFs  \\
		SprottTorus             & 1.8440  & 13.5801  & 2.0916  & 11.7551  & 0.0062  & 0.0174  & 0.0068  & 0.0262 & Regular KFs \\
		StickSlipOscillator     & 0.0149  & 0.0223   & 0.1144  & 0.0647   & 0.0060  & 0.0048  & 0.0025  & 0.0032 & Sparse KFs  \\
		SwingingAtwood          & 1.3206  & 5.9341   & 0.2365  & 0.4268   & 0.1911  & 0.9231  & 0.0878  & 0.3789 & Sparse KFs  \\
		Thomas                  & 0.0964  & 0.0221   & 0.1858  & 0.0187   & 0.6922  & 0.1694  & 0.0188  & 0.0074 & Trained RBF \\
		ThomasLabyrinth         & 1.1063  & 2.0744   & 1.4589  & 0.1824   & 2.3277  & 0.3354  & 0.5611  & 0.2008 & Trained RBF \\
		Torus                   & 0.0125  & 0.0823   & 0.0282  & 0.1320   & 0.0497  & 0.3391  & 0.0037  & 0.0104 & Sparse KFs  \\
		Tsucs2                  & 0.8522  & 4.5254   & 0.1237  & 0.1248   & 0.0064  & 0.0021  & 0.0276  & 0.0051 & Sparse KFs  \\
		TurchinHanski           & 0.2858  & 1.9826   & 0.1663  & 0.0357   & 0.0299  & 0.0195  & 0.0247  & 0.0178 & Sparse KFs  \\
		VallisElNino            & 0.0407  & 0.0839   & 0.0703  & 0.0197   & 0.0408  & 0.0258  & 0.0118  & 0.0037 & Sparse KFs  \\
		VossDelay               & 37.9067 & 19.0553  & 38.4574 & 18.7339  & 4.5635  & 12.6397 & 3.1540  & 3.3905 & Sparse KFs  \\
		WangSun                 & 1.9833  & 10.1195  & 0.5392  & 4.1739   & 1.9652  & 1.4283  & 0.0394  & 0.0252 & Sparse KFs  \\
		WindmiReduced           & 0.0584  & 0.8784   & 0.1360  & 1.9927   & 0.0158  & 0.6915  & 0.0096  & 0.2036 & Sparse KFs  \\
		YuWang                  & 0.3932  & 5.6805   & 0.1470  & 0.2048   & 0.0371  & 0.1076  & 0.1190  & 0.0828 & Sparse KFs  \\
		YuWang2                 & 0.0360  & 0.1737   & 0.0490  & 0.0474   & 0.0078  & 0.0256  & 0.0051  & 0.0128 & Sparse KFs  \\
		ZhouChen                & 4.5703  & 25.6922  & 3.2739  & 18.4165  & 0.2694  & 0.0129  & 0.2194  & 0.0134 & Sparse KFs       \\
		
	\end{supertabular}
\end{center}\label{tb}

\section{Conclusion}
In this work, we present a Sparse Kernel Flows algorithm for learning the 'best' kernel from data. The proposal provides an effective modification of the method of Regular Kernel Flows that allows sparsifying the initial base kernel and improves the accuracy. Our numerical experiments on a benchmark of 132 chaotic systems demonstrate that the learned kernel using Sparse Kernel Flows is able to accurately represent the underlying dynamical system responsible for generating the time series data.


\section*{Acknowledgment}
This research was completed during the first author's visit to Imperial College London; L.Y. thanks Boumediene Hamzi and Jeroen Lamb for hosting the visit. L.Y. thanks the Nanjing University of Aeronautics and Astronautics for funding through Interdisciplinary Innovation Fund for Doctoral Students KXKCXJJ202208. Naiming Xie acknowledges support from the National Natural Science Foundation of China (72171116), and the Fundamental Research Funds for the Central Universities of China (NP2020022).  HO acknowledges support from the Air Force Office of Scientific Research under MURI award number FA9550-20-1-0358 (Machine Learning and Physics-Based Modeling and Simulation) and the Department of Energy under the MMICCs SEA-CROGS award.  BH acknowledges support from the Air Force Office of Scientific Research (award number  FA9550-21-1-0317) and the Department of Energy (award number SA22-0052-S001).

\appendix
\section{Appendix}

\subsection{Reproducing Kernel Hilbert Spaces (RKHS)}

We give a brief overview of reproducing kernel Hilbert spaces as used in statistical learning
theory ~\cite{CuckerandSmale}. Early work developing
the theory of RKHS was undertaken by N. Aronszajn~\cite{aronszajn50reproducing}.

\begin{definition} Let  ${\mathcal H}$  be a Hilbert space of functions on a set ${\mathcal X}$.
Denote by $\langle f, g \rangle$ the inner product on ${\mathcal H}$   and let $\|f\|= \langle f, f \rangle^{1/2}$
be the norm in ${\mathcal H}$, for $f$ and $g \in {\mathcal H}$. We say that ${\mathcal H}$ is a reproducing kernel
Hilbert space (RKHS) if there exists a function $k:{\mathcal X} \times {\mathcal X} \rightarrow \RR$
such that\\
\begin{itemize}
 \item[i.] $k_x:=k(x,\cdot)\in{\mathcal{H}}$ for all $x\in{\mathcal{H}}$.
\item[ii.] $k$ spans ${\mathcal H}$: ${\mathcal H}=\overline{\mbox{span}\{k_x~|~x \in {\mathcal X}\}}$.
 \item[iii.] $k$ has the {\em reproducing property}:
$\forall f \in {\mathcal H}$, $f(x)=\langle f,k_x \rangle$.
\end{itemize}
$k$ will be called a reproducing kernel of ${\mathcal H}$. ${\mathcal H}_k$  will denote the RKHS ${\mathcal H}$
with reproducing kernel $k$ where it is convenient to explicitly note this dependence.
\end{definition}

The important properties of reproducing kernels are summarized in the following proposition.
\begin{proposition}\label{prop1} If $k$ is a reproducing kernel of a Hilbert space ${\mathcal H}$, then\\
\begin{itemize}
\item[i.] $k(x,y)$ is unique.
\item[ii.]  $\forall x,y \in {\mathcal X}$, $k(x,y)=k(y,x)$ (symmetry).
\item[iii.] $\sum_{i,j=1}^q\alpha_i\alpha_j k(x_i,x_j) \ge 0$ for $\alpha_i \in \RR$, $x_i \in {\mathcal X}$ and $q\in\mathbb{N}_+$
(positive definiteness).
\item[iv.] $\langle k(x,\cdot),k(y,\cdot) \rangle=K(x,y)$.
\end{itemize}
\end{proposition}
Common examples of reproducing kernels defined on a compact domain $\mathcal{X} \subset \mathrm{R}^n$ are the 
(1) constant kernel: $K(x,y)= m > 0$
(2) linear kernel: $k(x,y)=x\cdot y$
(3) polynomial kernel: $k(x,y)=(1+x\cdot y)^d$ for $d \in \N_+$
(4) Laplace kernel: $k(x,y)=e^{-||x-y||_2/\sigma^2}$, with $\sigma >0$
(5)  Gaussian kernel: $k(x,y)=e^{-||x-y||^2_2/\sigma^2}$, with $\sigma >0$
(6) triangular kernel: $k(x,y)=\max \{0,1-\frac{||x-y||_2^2}{\sigma} \}$, with $\sigma >0$.
(7) locally periodic kernel: $k(x,y)=\sigma^2 e^{-2 \frac{ \sin^2(\pi ||x-y||_2/p)}{\ell^2}}e^{-\frac{||x-y||_2^2}{2 \ell^2}}$, with $\sigma, \ell, p >0$.

\begin{theorem} \label{thm1}
Let $k:{\mathcal X} \times {\mathcal X} \rightarrow \RR$ be a symmetric and positive definite function. Then there
exists a Hilbert space of functions ${\mathcal H}$ defined on ${\mathcal X}$   admitting $k$ as a reproducing Kernel.
Conversely, let  ${\mathcal H}$ be a Hilbert space of functions $f: {\mathcal X} \rightarrow \RR$ satisfying
$\forall x \in {\mathcal X}, \exists \kappa_x>0,$ such that $|f(x)| \le \kappa_x \|f\|_{\mathcal H},
\quad \forall f \in {\mathcal H}. $
Then ${\mathcal H}$ has a reproducing kernel $k$.
\end{theorem}

\begin{theorem}\label{thm4}
 Let $k(x,y)$ be a positive definite kernel on a compact domain or a manifold $X$. Then there exists a Hilbert
space $\mathcal{F}$  and a function $\Phi: X \rightarrow \mathcal{F}$ such that
$$k(x,y)= \langle \Phi(x), \Phi(y) \rangle_{\mathcal{F}} \quad \mbox{for} \quad x,y \in X.$$
 $\Phi$ is called a feature map, and $\mathcal{F}$ a feature space\footnote{The dimension of the feature space can be infinite, for example in the case of the Gaussian kernel.}.
\end{theorem}

\section*{Data accessibility}
Data and codes are available at \url{https://github.com/Yanglu0319/Sparse_Kernel_Flows}.

\small\setlength{\bibsep}{5pt}
\bibliographystyle{cas-model2-names}
 \bibliography{reference,missing_dyn}

\end{document}